\documentclass[10pt,twocolumn,letterpaper]{article}

\usepackage{wacv}
\usepackage{times}
\usepackage{epsfig}
\usepackage{graphicx}
\usepackage{amsmath}
\usepackage{amssymb}
\usepackage{multirow}

\def\wacvPaperID{747} 

\wacvfinalcopy 

\ifwacvfinal
\def\assignedStartPage{9876} 
\fi

\ifwacvfinal
\usepackage[breaklinks=true,bookmarks=false]{hyperref}
\else
\usepackage[pagebackref=true,breaklinks=true,colorlinks,bookmarks=false]{hyperref}
\fi

\ifwacvfinal
\else
\fi

\begin{document}


\title{A Fast Partial Video Copy Detection Using KNN and Global Feature Database}

\author{Weijun Tan$^{1,2}$, Hongwei Guo$^2$, Rushuai Liu$^2$\\
$^1$LinkSprite Technologies, Longmont, CO 80501, USA\\
$^2$Shenzhen Deepcam Information Technologies, Shenzhen, China\\
{\tt\small weijun.tan@linksprite.com,\{hongwei.guo,rushuai.liu\}@deepcam.com}
}

\maketitle

\thispagestyle{empty}
\pagestyle{empty}

\begin{abstract}

Unlike in most previous partial video copy detection (PVCD) algorithms where reference videos are scanned one by one, we treat the PVCD as a video search/retrieval problem. We propose a fast partial video copy detection framework in this paper.  In this framework, all frame CNN features of the reference videos are organized in a KNN searchable database. Instead of scanning all reference videos, the query video segment does a fast KNN search in the global feature database. The returned results are used to generate a shortlist of candidate videos. A modified temporal network is then used to localize the copy segment in the candidate videos. Furthermore, We propose to use a transformer encoder to improve the CNN feature. We evaluate our algorithm on the VCDB dataset. Our benchmark F1 scores exceed the state of the art by a big margin. The speed of our algorithm is also improved significantly. 

\end{abstract}





\section{Introduction}

With the popularity of high-speed internet and 5G mobile communication, low-cost video capture devices, and social media, a large number of videos are generated and uploaded online every second.  This brings many issues including copyright, privacy, and child protection.  In this work, we focus on partial video copy detection (PVCD), which identifies videos and localizes copy segments in a large database of long untrimmed reference videos. 

The PVCD is a very challenging problem because of a large variety of editing and transformations such as rotation, scaling, cropping, captioning, watermarking, FPS change, picture in picture.  For the background, the readers are referred to \cite{VCDB2},\cite{TEMPORAL-NETWORK} and references therein. 

There are three main parts to the PVCD algorithm. The first one is to identify the videos which have copy segments in them. The second one is to represent a video with a feature. The last one is to localize the copy segment in a candidate video. We explore PVCD in all these three directions and our contributions are summarized as follows. 

Traditional PVCD algorithms using hand-crafted features can use features saved in an inverted file and use a fast search such as k-nearest-neighbor (KNN) approaches \cite{VCDB1, MYLiu, r2,r13}. However, since the deep CNN features are explored for PVCD, there are no works studying to use KNN approaches likely because the deep CNN features are hard to encode or compress further, and require a lot of resources in computation and storage. Instead, reference videos are scanned one by one \cite{VCDB2}.  

Our first contribution is to propose a PVCD framework to use fast KNN search on a database of deep CNN features. To our knowledge, this is the first work that uses a KNN search method on deep CNN features for PVCD.  We first construct a global feature database consisting of all frame features of all reference videos. Given a query segment, we use the frame features of the query segment to search in the global database and return the top-K answers for every query frame.  These answers' scores are accumulated by their reference video ID.  The reference videos whose scores are the highest are returned as the candidate videos.  

Our second contribution is to propose to use a transformer \cite{Transformer} encoder to improve the frame CNN feature. In \cite{VCDB2} and \cite{LAMV}, CNN is used to extract the feature for a video frame. This feature is for one individual frame image without considering the temporal context. Inspired by the transformer \cite{Transformer} for its power to model the context in the language model, image classification \cite{VIT}, object detection \cite{DETR}, video classification \cite{ViViT}, video action recognition \cite{Transformer-action-recognition}, video re-localization \cite{Transformer-VideoReloc}, video point cloud \cite{Fan_2021_CVPR} etc., particularly by \cite{VISIL-Ext}, we explore its application in an location-sensitive video representation for the PVCD. 

Our last contribution is on the copy segment localization algorithm. We still use the temporal network algorithm \cite{TEMPORAL-NETWORK}, but we make some major modifications to improve the performance as well as the speed. Since the feature and the similarity matrix are optimized, we can tighten the parameters such that the algorithm runs a lot faster while achieving the same or better performance.  

Please note that, our contributions are not about exploring brand-new ideas. Instead, we optimize existing techniques into a well-engineered framework to solve a practical problem. We evaluate the new PVCD framework on the VCDB dataset \cite{VCDB1}. The best F1 score is improved from the state-of-the-art 0.8025 to 0.8764 by a large margin. We also evaluate the speed in a few critical parts of the PVCD and show the expedition of PVCD. 

\section{Related Work}

Video copy detection includes near duplicate video detection or retrieval (NDVD or NDVR), PVCD, and content-based copy detection (CCD). For an example of NDVR, please refer to \cite{kordopatis2017dml} and references therein. For CCD, please refer to the TRECVID 2011 report \cite{TRECVID2011}. 

Two important parts of PVCD are the video representation and the copy localization. In early times, a global feature is generated to represent a video for its simplicity, fast speed, and low data storage \cite{r11,r3,r4}. However, since its performance is poor, we only consider PVCD using frame features. Traditional local features for PVCD include LBP, SIFT, and variants, which are then aggregated into frame features using BOW, Fisher vectors, or similar methods \cite{r2,r12,r13,TEMPORAL-NETWORK}. 

Before CNN feature is used, Hamming or binary embedding and some simple local sensitivity hashing methods are used \cite{VCDB1} on these handcrafted features. Other works using compact feature embedding include \cite{MYLiu,r2,r13}. In \cite{MYLiu}, a video fingerprint-based inverted search is implemented. In \cite{r2}, a multiple level search in inverted files is  implemented on a database of a single reference video.  Due to the significant inferior performance of these hard crafted features, we only consider CNN feature in this work. 

The work \cite{VCDB2} is the first one that explores deep CNN features on PVCD.  Since then, the LAMV \cite{LAMV} implements the PVCD in the frequency domain using the temporal matching kernel \cite{TMK}. Since CNN is used in PVCD and similar tasks \cite{VCDB2}, \cite{VISIL}, \cite{VISIL-Ext}, the reference videos are scanned one by one, and no fast KNN search approaches are available except for \cite{MMTA}. In \cite{MMTA}, a low-cost global descriptor in combination with a decision strategy adapted from reinforcement learning is used. 

For the copy localization, early PVCD works include \cite{HOUGH-VOTE}, and \cite{TEMPORAL-NETWORK}, and the references therein. In these two works, the authors propose the Hough voting algorithm and the temporal network (or network flow) algorithm, which find the location of the copy in the similarity matrix between the query video and the reference video. These two algorithms are evaluated in \cite{VCDB2} on CNN features. 

All these algorithms use the similarity matrix directly. For PVCD, this similarity matrix is ideally a diagonal sub-matrix at the location of the copy segment. In reality, however, it is not. The ViSiL \cite{VISIL} first explores the information hidden in the similarity matrix. The authors propose a simple CNN to train the noisy sub-matrix to a better diagonal sub-matrix for similar tasks such as NDVR, event video retrieval. In these tasks, they only care if there is a same event in the reference video, but do not consider the event's location. Another disadvantage is it has to scan every reference video to find the candidate video, which cannot be used for a real video retrieval problem.

In 2019, there was a PVCD challenge on Data Fountain organized by China Computer Federation \cite{BDCI-CCF}. In this challenge, the query was an entire video, instead of a video segment as in the VCDB standard protocol. In one of the forum discussions, one team used a KNN search in a global feature database, but the localization part was unclear \cite{HYData}. Part of our work is inspired by this scheme. Based on the KNN search in the global feature database, we develop a whole framework for PVCD.   

\section{Proposed Method}

\begin{figure}[t]
    \centering
    \includegraphics[scale=0.7]{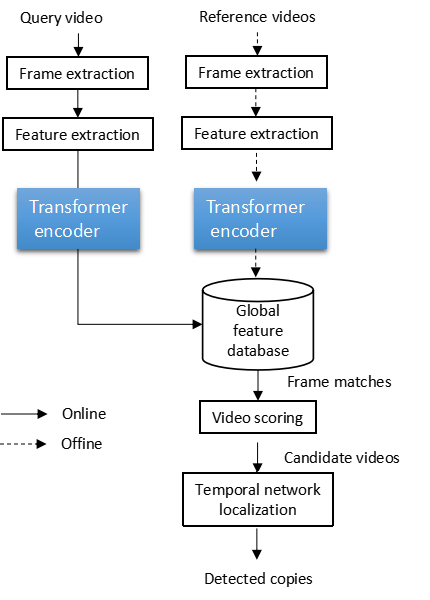}
    \caption{Diagram of our PVCD algorithm}
    \label{fig1}
\end{figure}

\subsection{System Diagram}

The diagram of our PVCD algorithm is shown in Figure \ref{fig1}. There are three main functions. The first one is the video representation. In this work, the frame is extracted in the video uniformly Then; the CNN feature is generated for the frame. This feature is for a single frame without considering the temporal context. To overcome this deficiency, we propose to use a transformer encoder, highlighted in blue in Figure \ref{fig1}, to improve the representation of the video, which we call representation learning in this paper.             

The second function is the global feature database - the foundation for a fast KNN search and video scoring. The output of this function is a shortlist of candidate videos that have the closest matches to the query video. More details will be given shortly in Section 3.2.    
The third function is the copy segment localization, which finds the location of the copy segment in the candidate videos. We propose to use a diagonalization CNN, highlighted in green in Figure 2, to refine the similarity matrix. After that, the temporal network algorithm is used to localize the copy segment.  The outputs are the starting and ending location of the copy segments and their similarity scores.    

Please note that there are offline and online processes in this diagram, denoted with solid and dashed lines in Figure \ref{fig1}. In the offline process,  the global feature database is created from all the reference videos. This offline process does not affect the online query speed. In the online process, the query video is processed to generate frame features, then the KNN search and copy detection. Please note that this query approach does not scan videos. It first finds candidate videos from the KNN search results. Then copy detection is done on this shortlist of candidate videos. 

\subsection{Global Feature Database}

In this algorithm, frames are uniformly extracted out of every reference video at one frame per second (FPS=1).  A CNN feature is generated for every frame. These features are all inserted into the global feature database. The video ID and frame index are combined to index the global feature database.  

Assume there are $N$ reference videos, denoted by $r_i,i=1,2,...,N$. Let the extract frames of the video $r_i$ be $r_{i,j}$, where $j=1,2,...,M_i$, and $M_i$ is the number of extracted frames in video $r_i$. For simplicity, we also use $r_{i,j}$ to represent the CNN feature of this frame, without ambiguity. So the features of the reference video $r_i$ can be described with a matrix $R_i = [r_{i,1}; r_{i,2}; ... ; r_{i,M_i}]$, which has a dimension $M_i \times d$, where $d$ is the length of the feature.  The global feature database for all the reference videos consists of features ${r_{i,j}}$, where $i=1,2,...,N$, $j=1,2,...,M_i$ in ascending order on $j$ then on $i$. The index $i$ is mapped to a video ID if a video ID instead of number index is preferred. The index for the database is a composite index of the video ID $i$ and frame index $j$. This function is pre-processed on the reference video database and is indicated as offline in Figure 2.   
Since all CNN features are $L_2$ normalized, the cosine distance, a measure of the similarity of a pair of frames, is equivalent to the inner product or Euclidean distance. These two distances are the most commonly used metric in KNN algorithms. So this global feature database is KNN-searchable with any KNN algorithm.        

Similarly, assume $q_t, t=1,2,...,T$ be the frame feature of the query video, where $T$ is the number of extracted frames. The feature matrix is $Q = [q_1; q_2; ... ; q_M]$. When this query video detects a copy in the reference video database, the frames and features are generated in the same way on the query video, and the features are used to search in the global feature database.  This is indicated as online process in Figure 2.  For every frame $q_t, t=1,2,...,M$, the first top-K (denoted $top\_K\_all$) matches  $r^t_{vid(l),j(l)}$ are returned, where $l=1,2,..,K$, $vid(l)$ is the video ID of the $l-th$ match, and the $j(l)$ is the frame index of this match. In other words, 
\begin{multline}
    sim(r^t_{vid(1),j(1)})>=sim(r^t_{vid(2),j(2)})>=\\
    sim(r^t_{vid(3),j(3)})>=...>=sim(r^t_{vid(K),j(K)})
\label{EQ1}
\end{multline}
where $sim()$ is the frame similarity score. 

Next in the video scoring function, every returned answer is assigned to a video according to answer's video ID as follows,  

\begin{equation}
sim(vid) = \sum_{t} \sum_{l} sim(r^t_{vid(l),j(l)}) \; s.t. \;  vid(l)=vid \end{equation}

The videos whose total similarity scores $sim(vid)$ are the highest are the candidate videos.  We return the top-K (denoted $top\_K\_video$) videos for every query, so the returned candidate videos $vid_i$ have, 

\begin{equation}
sim(vid_1)>= sim(vid_2)>=...>=sim(vid_K)
\end{equation}
The value of $top\_K\_video$ can be chosen appropriately for a particular dataset. 

\subsection{Modified Temporal Network Localization}

After the candidate videos are available, the temporal network algorithm \cite{TEMPORAL-NETWORK} is used to localize the position of the copy in the candidate videos.  Since the number of candidate videos is usually very small, the running speed is acceptable, close to real-time.   

There are a few key parameters that affect the temporal network localization performance. The first one is the top-K matched (denoted $top\_K\_one$) frames for every frame in the query. Only the top K frames with the highest matching scores are used, and the rest are ignored. The second parameter is the $max\_step$, which is the allowed maximum step in both row and column directions in the similarity matrix when a network flow path is formed, which is the same as the constraint $wnd$ in \cite{TEMPORAL-NETWORK}. The third one is a similarity threshold $sim_{th}$ where similarities smaller than this threshold are not considered.  

Consider the similarity matrix between a pair of videos ($q$,$r$), where $q$ is the query video segment, and $r$ is one reference video found in the previous step. Let us call every element in the matrix a 2D point $p$, where $p[0]$ and $p[1]$ are its row and column index. The similarity matrix is expressed as  $sim(p)=sim(q_{p[0]},r_{p[1]})$. Please note that for simplicity, we drop the video index of $q$ and $r$ since there is only one $q$ and one $r$ in this pair. The index $i$ is now the frame index.  The temporal network localization becomes a network flow optimization problem, 

\begin{multline}
path(p),score,L = \max_p {{\sum_{i}^L{sim(p_i)}}} \; s.t. \; \\ 
(a): 0<p_{i+1}[0]-p_{i}[0]<=max\_step    \\
(b): 0<p_{i+1}[1]-p_{i}[1]<=max\_step    \\
(c): p_i[1] \in top-K(p_i[0])   \\
(d): sim(p_i)>= sim_{th}   \\
(e): |p_{i+1}[0]-p_{i}[0] - (p_{i+1}[1]-p_{i}[1])| < max\_diff \\
\end{multline}

The conditions (a) and (b) are in the original temporal network algorithm \cite{TEMPORAL-NETWORK}.  The conditions (c)-(e) are our modifications.  These modifications show a big impact on both speed and performance. In the original temporal network \cite{TEMPORAL-NETWORK}, the network flow runs in the whole similarity matrix. In conditions (c) and (d), we propose to only keep the reliable top-K matched reference video frames for every query frame. As a result, the similarity matrix becomes a sparse matrix, and the network flow runs more accurately and efficiently.  

The condition (e) is added to force the network flow to run along with a narrow band along the diagonal direction.  Since we optimize the similarity matrix to near a diagonal matrix in the copy segment location, we can tighten the conditions by using smaller $K$ and $max\_diff$ and expedite the algorithm.  

The returned result includes the longest $path$ of length $L$ whose total $score$ is maximized. From the path ${p_1,p_2,...,p_L}$, the starting and ending location of copy clip in $q$ and $r$ are defined, as well as the video similarity score, 

\begin{equation}
\begin{aligned}
& q_{start} = p_1[0],\;  q_{end} = p_L[0] \\ 
& r_{start} = p_1[1],\;  r_{end} = p_L[L] \\
& sim(q,r) = score/L \\
\end{aligned}
\end{equation}

\subsection{CNN Feature with Transformer}

The transformer is introduced in the paper \cite{Transformer}. It is one of the breakthroughs in machine learning in recent years. It has been studied extensively on different applications, including \cite{VIT, DETR, ViViT,Transformer-action-recognition, Transformer-VideoReloc}, etc. Particularly in \cite{VISIL-Ext}, transformer is proposed to improve representation CNN feature in a video retrieval task. Therefore, we explore its application in PVCD - a video retrieval and re-localization task, where we add the supervision of location.   

The self-attention mechanism of the transformer is effective at modeling long-term dependencies within a sequence input. The encoder can be used to aggregate temporal context in representing videos. Even though the encoder output keeps the same dimension as the input, the contextual information within a longer range of each frame feature is accounted for.  

\begin{figure}[t]
    \centering
    \includegraphics[scale=0.5]{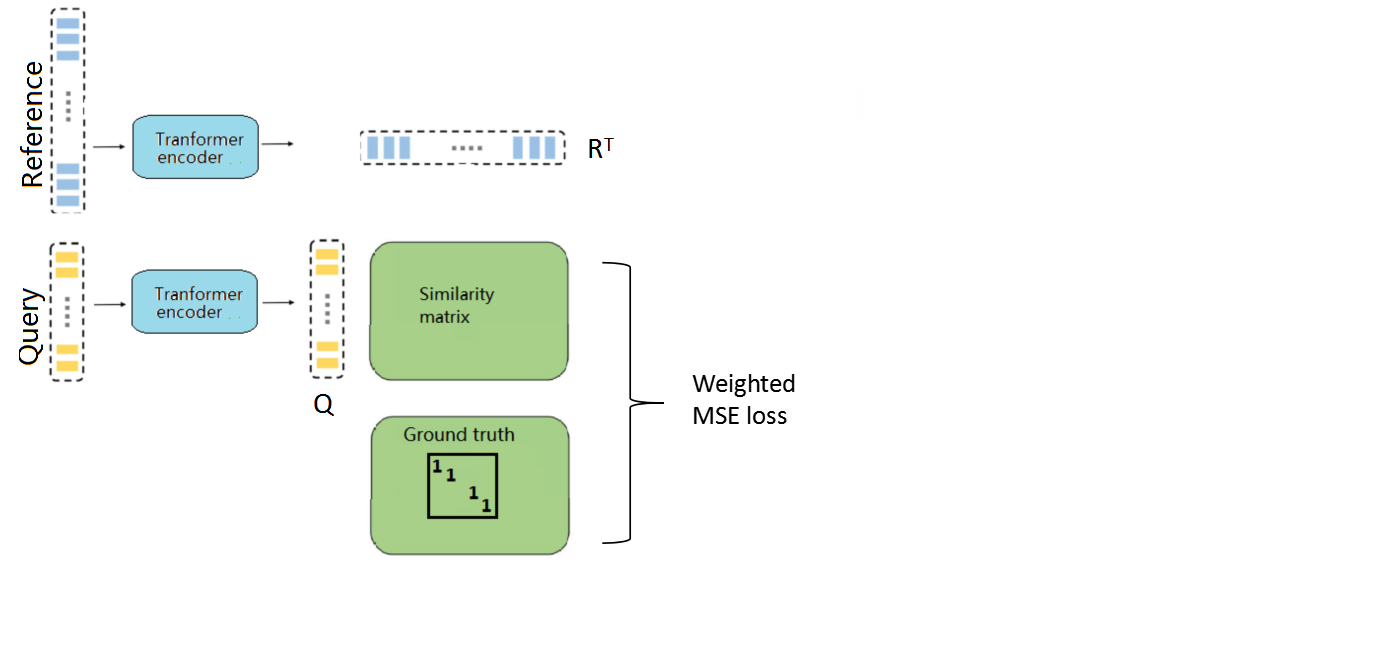}
    \caption{CNN feature using a transformer encoder.}
    \label{fig2}
\end{figure}

Let the video descriptor $X$ of dimension ${M \times d}$.  Denote the parameter matrices of the transformer encoder by $W_Q$,$W_K$, $W_V$. The video descriptor X is encoded into Query Q (please do not confuse with the query video feature matrix), Key K and Value V by three different linear transformations, 

\begin{equation}
Q = X^TW_Q, K = X^TW_K, V = X^TW_V
\end{equation}
The self-attention is calculated as,

\begin{equation}
Att(Q, K, V ) = softmax ( \, \frac{QK^T}{\sqrt{d_k}} ) \,V
\end{equation}
where $\sqrt{d_k}$ is used to scale the dot product. This self-attention value is then passed to a LayerNorm layer and a Feed-Forward Layer to get the output of the Transformer encoder. The multi-head attention is usually used, where the multi-head attention outputs are concatenated before the Feed-Forward layer.    
This transformer encoder can be used on different video frame features. In this work, it is applied to the Resnet-RMAC \cite{LAMV} feature. Our proposed transformer encoder for the CNN feature is shown in Figure \ref{fig2}. 

\begin{figure}[t]
    \centering
    \includegraphics[scale=1.1]{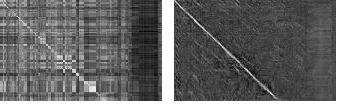}
    \caption{An example of an original similarity matrix and its transformer encoded one.}
    \label{fig3}
\end{figure}

An example of a similarity matrix before and after applying the transformer is shown in Figure \ref{fig3}. In this example, it can be clearly seen that after the transformer encoder, the similarity matrix becomes more like a diagonal sub-matrix at the copy segment location. 

\subsection{Loss Function}

We use the notation defined in Section 3.1. For query video feature matrix $Q$ of dimension $T \times d$, and the reference video feature matrix $R$ (we omit index $i$ for simplicity) of dimension $M \times d$, the similarity matrix $P$ is,
\begin{equation}
P =[P_{i,j}] = QR^T
\end{equation}

Assume there is a copy segment of the query video Q in the reference video R. Let the starting location and length of the copy segment in the query video be $(s_q,L)$, in the reference video be $(s_r,L)$. Then the ground-truth (GT) similarity matrix $P_{GT}$ is, 

\begin{equation}
\begin{aligned}
&P_{GT} = [P'_{i,j}]\\
&P'_{s_q+i,s_r+i} = 1,i = 0,1,...,L-1\\
&P'_{i,j} =0 \; elsewhere
\end{aligned}
\end{equation}

The loss function is defined as the mean square error (MSE) of all the elements between matrix $P$ and $P_{GT}$, 

\begin{equation}
loss = \frac{1}{MT} \sum_{i,j}{[(P_{i,j}-P'{i,j})w_{i,j}]^2}
\end{equation}
where the $w_{i,j}$ is a weight factor to balance the number of ones and number of zeros in the similarity matrix, since the number of zeros is a lot more than the number of ones, we use a weight of 0.1 for zeros and 1.1 for ones, as an example.    

Please note that there are some arguments that PVCD may not have an identity sub-matrix at the copy segment location due to video rate (FPS) change or reshuffling. This is partially true. However, in PVCD, we use the segment F1 as a metric and treat detection of any frames as a successful detection of the copy segment. Furthermore, since adjacent frames are usually very similar, an identity submatrix is likely showing up at some locations of the copy segment. The evaluation results on the VCDB dataset will demonstrate our algorithm's effectiveness.   

\section{Experiments}

\subsection{Implementation Details}

All the videos are first uniformly extracted with 1 FPS into frames, as widely used in many works including \cite{VCDB1,VCDB2,HengYang}. We test other FPS and find 1 FPS gives a good trade-off between performance and speed. When the scale of the reference videos gets very large, larger FPS can be used to speed up the generation of the CNN features. The impact is that the resolution of the PVCD is reduced. 

Then the Resnet29-RMAC CNN feature \cite{LAMV} is extracted for every frame. In one ablation study, we also extract the VGG16 max-pooling feature. The dimension of feature $d=512$. 

We use the global frame feature database and the temporal network for copy segment localization. The parameters we use are ($top\_K\_all$, $top\_K\_one$, $top\_K\_video$, $sim_{th}$, $max\_step$, $max\_diff$), whose meanings are defined in Sections 3.1 and 3.2. For this study, we do not use the KNN search method. Instead, matrix multiplication and sorting are used to return the top-K matches.  

In the training of the transformer encoder, we use a multi-head transformer with inner dimension 128. We use the Adam optimizer, learning rate 1E-6 and 30 epochs. The weight factor 0.1 and 1.1 are used on the ones and zeros elements in the similarity matrix.    

In the training of the diagonalization CNN, we use the Adam optimizer, learning rate 1E-4, and 30 epochs. No weight factors are used on the ones and zeros elements in the similarity matrix.    

For the speed test, we use an Intel i7-6850K CPU at 3.60GHz and a Ubuntu 16.04 LTS OS. We use an Nvidia GTX 1080TI GPU in the GPU mode of the KNN speed test. 

\subsection{Dataset}

 VCDB \cite{VCDB1} is the only public PVCD dataset.  VCDB includes 528 videos in the core dataset and a very large background dataset. There are a total of 9236 annotated copy segments. In this section, we evaluate our algorithm on the VCDB dataset. We only use the videos in the core dataset and use all the annotated copy segments as the query. As in the standard VCDB protocol, the detected segment with any frame overlap with the ground truth is called a true positive. We only use segment F1 score in this study. One frame per second is extracted uniformly in all videos. 
 
 However, for training the transformer, we cannot use the same VCDB dataset for both training and test since the whole VCDB dataset is used in the benchmark performance test \cite{VCDB1,VCDB2,LAMV}. Instead, we use a private PVCD dataset as a training dataset and use the whole VCDB dataset as a test dataset. This makes the benchmark on the VCDB dataset more reasonable and also demonstrates this algorithm's cross-domain performance. This dataset has more than 300 reference videos, most of which are longer than the VCDB reference videos. There are a few thousand query videos, each of which has a copy segment in multiple reference videos.     

\subsection{Training Data Collection for Transformer}

In ViSiL \cite{VISIL}, \cite{VISIL-Ext}, and \cite{Transformer-VideoReloc}, triplet loss or contrastive learning is used, therefore triplet videos including anchor video, positive video, and negative video are prepared. In our work, there is no triplet video needed. We use the standard supervised learning using training samples (positive or negative) and their ground truth. So the collection of training data is easier. 

The positive samples are prepared from the annotations of the training dataset. We borrow the idea of using hard negative samples in ViSiL \cite{VISIL}. We first do the PVCD on the features before the transformer is applied, collect all the false-positive results as the hard negative samples. If the numbers are not enough, we can collect more negative samples by randomly choosing video segments in the reference videos in the training dataset. This has been proved much better than only randomly choosing video segments in the reference videos. 

Furthermore, since the number of negative samples is a lot more than the positive samples, we randomly pick negative samples from its pool in every epoch of the training process and make the numbers of positive and negative video segments equal in any epoch.   

\subsection{KNN Search in Global Feature Database}

In the following subsection, we study the global feature database and the CNN feature without using the transformer or the similarity CNN. 

After the candidate video is found, there are two methods to form the similarity matrix.  We can use use the original similarity matrix calculated from every pair-wise frame between the query segment and the candidate video, or we can use the reconstructed similarity matrix using the returned KNN results. Only the frames found by the KNN belonging to this candidate video have a non-zero value in the similarity matrix; others are all filled with zero. As a result, this similarity matrix is very sparse, and the temporal network localization runs a lot faster than the original similarity matrix.  If the performance is comparable, then the reconstructed similarity matrix is preferred for its fast speed. 

Other parameters we try to optimize include $top\_K\_all$, and ($top\_K\_one$, $top\_K\_video$, $sim_{th}$, $max\_step$). To expedite the tuning, we only use the first 500 query segments.  Please note that, for a fair comparison, we must use the same set of query segments in this comparison. We should not compare these results with results for a different set of query segments.  The F1 score on the full query dataset may be lower than that on the first 500 segments. This is understandable because the distribution of the video similarity score of the first 500 segments is not stable yet.  

We first use the reconstructed similarity matrix to optimize parameters, and the results are listed in the first panel in Table \ref{T0}.  From the results, we see that $top\_K\_video$=20 have better results than $top\_K\_video$=10, because in VCDB dataset many segments have copy in more than 10 reference videos. The similarity threshold $sim_{th}$ = 0.5,0.6 give good results. The best result of F1=0.9010 is achieved with ($top\_K\_one$, $top\_K\_video$, $sim_{th}$, $max\_step$)=(20.20.0.5,5).  

Next, we check the performance of the original similarity matrix with a few good parameter configurations from the previous step, and the results are listed in the second panel in Table \ref{T0}. The best F1=0.8988 is achieved with $top\_K\_one$=20, $top\_K\_video$=20, $sim_{th}$=0.6, $max\_step$=5. From these results, we see that the reconstructed similarity matrix can achieve the same or even better performance than the original similarity matrix.   

Thirdly, we test scanning the reference videos. In this mode, the original similarity matrix is used, and $top\_K\_video$ is not used. The results denoted by S are shown in the third panel in Table \ref{T0}. We notice the big improvement from $sim_{th}$=0.0 to $sim_{th}$=0.5, which demonstrates the effectiveness of the modified temporal network localization algorithm. Furthermore, the speed of $sim_{th}$=0.5 is a lot ($>50$ times) faster than $sim_{th}=0.0$ or using no threshold in the original temporal network.  

\begin{table}
	\begin{center}
		\caption{Tuning parameters using the first 500 query segments. In the matrix column, R indicates reconstructed similarity matrix, O indicates original similarity matrix, and S indicates scanning videos. The parameters are  ($top\_K\_one$,$top\_K\_video$,$sim_{th}$,$max\_step$).} 
		\label{T0}
		\begin{tabular}{ccc}
		    \hline
			Matrix & Parameters & F1\\
			\hline
			R & (5, 10, 0, 10) & 0.7917 \\
			R & (10, 10, 0, 10) & 0.8015 \\
			R & (10, 20, 0, 10) & 0.8584 \\
			R & (20, 20, 0, 10) & 0.8609 \\
			R & (20, 20, 0, 5) & 0.8609 \\
			R & (20, 20, 0.25, 10) & 0.8637 \\
			R & (20, 20, 0.5, 10) & 0.8980 \\
			R & (20, 20, 0.5, 5) &  \textbf{0.9010} \\
			R & (20, 20, 0.6, 10) & 0.8959 \\
			\hline
			O & (20, 20, 0, 10) & 0.8774 \\
			O & (20, 20, 0.25, 10) & 0.8654 \\
			O & (20, 20, 0.5, 5) & 0.9010 \\
			O & (20, 20, 0.6, 10) &  0.8959\\
			O & (20, 20, 0.6, 5) &  \textbf{0.8988}\\
			\hline
			S& (20, - , 0.5, 10) &  0.7352\\
			S& (20, - , 0.0, 10) &  0.6694\\
			\hline
		\end{tabular}
	\end{center}
\end{table}

\subsection{Other CNN Features}

Other than the Resnet-29 RMAC feature from the LAMV \cite{LAMV}, we also use the VGG-16 pool5 feature map \cite{VGG16}. We do not directly use the FC output since it is widely believed that the Conv layer feature works better than the FC feature in the image retrieval task. We test the following pooling method: Max pooling, average pooling, Crop, SPoC, and find that the max-pooling gives the best F1 score. The results are summarized in Table \ref{T1}. For all these different pooling methods, please refer to \cite{Hu2020PyRetri} for details. The Resnet-29 RMAC F1 score outperforms the VGG-16 max-pooling by a big margin. 

We analyze the performance improvements in the F1 score. Using the better feature is a big factor in it. Changing the feature from VGG-16 max pooling to the Resnet-29 RMAC, the F1 score improves from 0.7657 to 0.8613.  However, with the same Resnet-29 RMAC feature, in Table 2, the F1 score is only 0.7352 by scanning all reference videos with a $sim_{th}=0.5$. With $sim_{th}=0$, the F1 score is even worse 0.689. This shows the improvement due to the modified temporal network localization algorithm.  While with the global feature database, the F1 score improves to 0.8613. With all these factors added together, our PVCD framework achieves significantly improved performance.      

\begin{table}
	\begin{center}
		\caption{Different CNN features and pooling methods.}
		\label{T1}
		\begin{tabular}{cc}
		    \hline
			CNN Feature & F1 Score\\
			\hline
			AlexNet FC, scanning videos \cite{VCDB2} & 0.6503\\ 
			\hline
			VGG-16 Max pooling & 0.7657 \\
            VGG-16 Avg pooling & 0.7656 \\
            VGG-16 SPoC & 0.7538 \\
            VGG-16 Crow & 0.7538 \\
            Resnet-29 RMAC & \textbf{0.8613} \\
			\hline
		\end{tabular}
	\end{center}
\end{table}

\subsection{CNN Features with Transformer}

In this section, we add the transformer encoder on the Resnet-29 RMAC CNN feature. We study to determine if this extra transformer encoder can improve the F1 score further over the previous two subsections. We use the previous F1 = 0.8613 as a baseline.  

We test a 2-head and an 8-head transformer on the full VCDB query dataset. The only difference is that the original CNN feature in Figure \ref{fig1} is replaced with the transformer encoded feature. The F1 score results are listed in Table \ref{T2}, as well as the parameters used to achieve this F1 score. Please note that the speed does not include the extraction of frames from videos and the extraction of the original CNN features from frames, which are pre-processed and saved. 

\begin{table}
	\begin{center}
		\caption{F1 scores and speed test results of transformer (TF) encoded feature on the VCDB query dataset. The parameters are $top\_K\_all$=200, ($top\_K\_one$, $top\_K\_video$, $sim_{th}$, $max\_step$, $max\_diff$). Transformers with 2-head (2H) and 8-head (8H) are tested. } 
		\label{T2}
		\begin{tabular}{cccc}
		    \hline
			Method & Parameters & F1 score & speedup\\
			\hline
            Baseline & (20,20,0.5,5,5) & 0.8613 & 1\\
            2H TF & (3,20,0.55,2,2) & 0.8709 & 1.64 \\
            2H TF & (3,20,0.55,2,0) & 0.8714 & 1.68 \\
            8H TF & (3,20,0.55,2,2) & 0.8760 & 1.42 \\
            8H TF & (3,20,0.55,2,0) & \textbf{0.8764} & 1.59 \\
            \hline
            VGG16-MP & (20,20,0.5,5,5) & 0.7657 & - \\
            VGG16-MP,2H & (20,20,0.5,5,5) & 0.8077 & - \\
			\hline
		\end{tabular}
	\end{center}
\end{table}

From the results, we notice that the transformer improves the F1 score from 0.8613 to 0.8760.  The improvement is not significant, perhaps because the previous baseline is already very good. In fact, the temporal network can model the temporal information in the copy localization algorithm indirectly, but it has to work on a pair of videos. The transformer, on the other hand, models the temporal information directly. The results show that the combination of the transformer and the temporal network localization algorithm brings some extra benefit, demonstrating that some temporal information is missed in the temporal network localization algorithm alone.       

However, if we pay attention to the parameters, we notice that a lot smaller parameters ($top\_K\_all$, $top\_K\_one$, $max\_step$, $max\_diff$) are used, which can help increase the speed. Please note that $max\_diff$ = 0 means the network flow is only allowed along the diagonal in the similarity matrix. It indicates that a simple dynamic programming algorithm can be used for copy segment localization. 

\textbf{Ablation study}: We use the 2-head and an 8-head transformer as an ablation study. The results show that a 2-head transformer is good enough. More heads get only a small performance gain while the model size gets larger quickly. We also test different $d_k$, inner dimensions and find that they have a very small impact on the performance.  

We also test the transformer encoder on the VGG16 max-pooling (denoted VGG16-MP in Table \ref{T2}) feature, and the results are listed in Table \ref{T2}. The original F1 score is 0.7657. After the transformer encoder, the F1 score increases to 0.8077. This more significant improvement shows the effectiveness of transformer encoder on CNN feature not as good as the Resnet29-RMAC feature \cite{LAMV}.  

\begin{figure*}[t]
    \centering
    \includegraphics[scale=0.4]{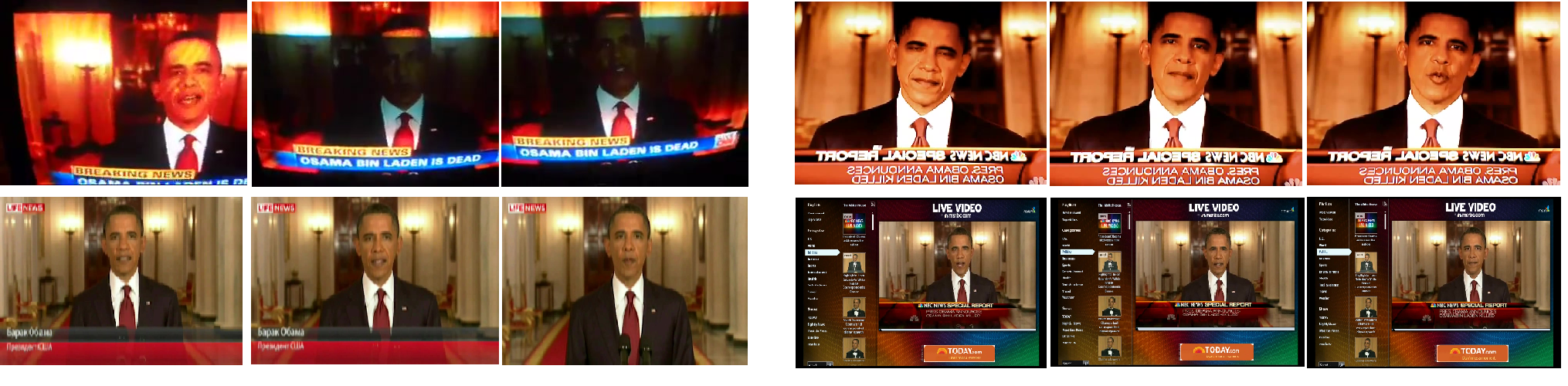}
    \caption{Examples, the first row is a successful detected case, the second row is a failed case.}
    \label{fig4}
\end{figure*}

\textbf{Qualitative example}: Shown in Figure \ref{fig4} are examples of a successful and a failed case. In the successful case in the first row, the video on the left is significantly shaded and distorted. The target video on the right is flipped but is still partially detected. In the failed case in the second row, the target video has a relatively small actual content window. This failed case indicates that there is still work to do in PVCD.         

\subsection{Comparison with the State of the Art}

We use our best F1 score on the full query dataset in Table \ref{T1} and Table \ref{T2} as our benchmark and compare it with the previous state-of-the-art results.  The comparison is listed in Table \ref{T3}. We notice that our CNN feature with transformer outperforms all existing methods. The F1 score improves from the previous 0.8025 to 0.8764. In addition, our algorithms improve the copy detection speed, as already shown in Table \ref{T2}, and more details of the speed test of KNN with GPU will be given in the next subsection.

\begin{table}
	\begin{center}
		\caption{Comparison with existing methods on VCDB dataset.}
		\label{T3}
		\begin{tabular}{cc}
		    \hline
			Method &  F1 Score\\
			\hline
			CNN \cite{VCDB2} & 0.6503 \\
			LAMV \cite{LAMV} & 0.6740 \\
			Ref \cite{MYLiu} & 0.6440 \\
			Ref \cite{MMTA} & 0.8025  \\
			Ours (No-transformer) & 0.8613  \\
			Ours (Transformer) & \textbf{0.8764}  \\
			\hline
		\end{tabular}
	\end{center}
\end{table}

\subsection{KNN Speed Test}

In this section, we test the query speed of different search methods.  The time reported here is only the query search time, not including the network flow localization time.

The baseline search method we test is a brute force search and sorting in matrix multiplication. Then we test the FAISS index types: Flat-CPU, Flat-GPU, IVF-Flat-GPU \cite{FAISS}. Since the total number of frames in the VCDB dataset is less than 100,000, we do not test the scalar or product quantization. The average search time per frame, the speedup over the matrix multiplication are summarized in Table \ref{T4}.  We see that the speedup of the Flat-CPU is about eight times, and that of Flat-GPU and IVF256-Flat-GPU 15 ad 11 times, respectively.   

\begin{table}
	\begin{center}
		\caption{Average search time per frame}
		\label{T4}
		\begin{tabular}{ccc}
		    \hline
			Method &  time(ms) & Speed up \\
			\hline
			Matrix multiplication & 20.0 & 1  \\
			Flat-CPU & 2.55 & 7.86 \\
			Flat-GPU & 1.48 & 14.79 \\
			IVF256-Flat-GPU & 1.82 & 10.81 \\
			\hline
		\end{tabular}
	\end{center}
\end{table}

\section{Conclusion}

In this work, we study a global feature database and a fast KNN search method for PVCD to find candidate videos that likely have copy segments in them. Using Resnet29-RMAC \cite{LAMV} feature, we improve the previous F1 score around 0.80 to 0.8613.  

Then we study an algorithm to improve the similarity of video pairs in PVCD. The transformer encoder enhances the representation of the frame CNN feature. The transformer improves the F1 score on the full VCDB query dataset further to 0.8764. 

Lastly, we study a modified temporal network to localize copy segments in the candidate videos. The modifications corporate with the improved representation or video similarity to improve both the F1 score and the speed.  

Overall, we improve the F1 score of PVCD on the VCDB dataset from the previous state-of-the-art 0.8025 to the new 0.8764.    

\bibliographystyle{ieee_fullname}
\bibliography{egbib}

\end{document}